\documentclass[11pt,a4paper]{article}
\usepackage[dvipsnames, table,xcdraw]{xcolor}
\usepackage[hyperref]{acl2019}
\usepackage{times}
\usepackage{latexsym}
\usepackage{multirow}
\usepackage{amsfonts}
\usepackage{graphicx}
\usepackage{amsmath}
\usepackage{footnote}
\usepackage{graphicx}
\usepackage{hyperref}
\hypersetup{breaklinks=true}
\usepackage{url}
\usepackage{soul}
\usepackage{booktabs}
\usepackage{enumitem}
\newcommand*{\red}{\textcolor{red}}
\newcommand*{\blue}{\textcolor{blue}}

\DeclareRobustCommand{\hlgreen}[1]{{\sethlcolor{LimeGreen}\hl{#1}}}
\DeclareRobustCommand{\hlmelon}[1]{{\sethlcolor{Melon}\hl{#1}}}
\DeclareRobustCommand{\hlyellow}[1]{{\sethlcolor{Yellow}\hl{#1}}}
\DeclareRobustCommand{\hlcyan}[1]{{\sethlcolor{Cyan}\hl{#1}}}

\aclfinalcopy 

\title{Revisiting Challenges in Data-to-Text Generation with Fact Grounding}

\author{Hongmin Wang \\
  University of California Santa Barbara \\
  \texttt{hongmin\_wang@cs.ucsb.edu}}

\date{}

\begin{document}
\sloppy
\maketitle


\begin{abstract}
Data-to-text generation models face challenges in ensuring data fidelity by referring to the correct input source.
To inspire studies in this area, \citet{ws17} introduced the {\tt RotoWire} corpus on generating NBA game summaries from the box- and line-score tables. 
However, limited attempts have been made in this direction and the challenges remain. 
We observe a prominent bottleneck in the corpus where only about 60\% of the summary contents can be grounded to the boxscore records. 
Such information deficiency tends to misguide a conditioned language model to produce unconditioned random facts and thus leads to factual hallucinations. 
In this work, we restore the information balance and revamp this task to focus on fact-grounded data-to-text generation.
We introduce a purified and larger-scale dataset, {\tt RotoWire-FG} (Fact-Grounding), with 50\% more data from the year 2017-19 and enriched input tables, hoping to attract more research focuses in this direction. 
Moreover, we achieve improved data fidelity over the state-of-the-art models by integrating a new form of table reconstruction as an auxiliary task to boost the generation quality.

\end{abstract}


\section{Introduction}
\label{sec:intro}

Data-to-text generation aims at automatically producing descriptive natural language texts to convey the messages embodied in structured data formats, such as database records~\citep{ChisholmRH17}, knowledge graphs~\citep{gardent-etal-2017-webnlg}, and tables~\citep{LebretGA16, ws17}. 
\autoref{tab:example} shows an example from the {\tt RotoWire}\footnote{\url{https://github.com/harvardnlp/boxscore-data}} ({\tt RW}) corpus illustrating the task of generating document-level NBA basketball game summaries from the large box- and line-score tables\footnote{Box- and line-score tables contain player and team statistics respectively. For simplicity, we call the combined input the \textit{boxscore table} unless otherwise specified.}.
It poses great challenges, requiring capabilities to select \textit{what to say} (content selection) from two levels: what entity and which attribute, and to determine \textit{how to say} on both discourse (content planning) and token (surface realization) levels. 

Although this excellent resource has received great research attention, very few works~\citep{Li018,ratishaaai, ratishacl, iso-etal-2019-learning} have attempted to tackle the challenges on ensuring data fidelity. This intrigues us to investigate the reason behind and we identify a major culprit undermining researchers' interests: the ungrounded contents in the human-written summaries impedes a model to learn to generate accurate fact-grounded statements and leads to possibly misleading evaluation results when the models are compared against each other.

\begin{table*}[]
\centering
\small
\setlength{\tabcolsep}{2pt}
\begin{tabular}{lcccccc}
\hline
TEAM & WIN & LOSS & PTS & FG\_PCT & BLK & ... \\\hline
Rockets & 18 & 5 & 108 & 44 & 7& \\\hline
Nuggets & 10 & 13 & 96 & 38 & 7& \\\hline\hline
PLAYER & H/A & PTS & RB & AST & MIN & ... \\\hline
James Harden & H & 24 & 10 & 10 & 38 & ... \\
Dwight Howard & H & 26 & 13 & 2 & 30 & ... \\
JJ Hickson & A & 14 & 10 & 2 & 22 & ... \\\hline
\multicolumn{7}{l}{\begin{tabular}[c]{@{}l@{}}\textbf{Column names} :\end{tabular}}\\
\multicolumn{7}{l}{\begin{tabular}[c]{@{}l@{}}H/A: home/away, PTS: points, RB: rebounds, \\ AST: assists, MIN: minutes, BLK: blocks, \\ FG\_PCT: field goals percentage\end{tabular}} \\
\\
\multicolumn{7}{l}{\begin{tabular}[c]{@{}l@{}}\textbf{An example hallucinated statement} :\end{tabular}}\\
\multicolumn{7}{l}{\begin{tabular}[c]{@{}l@{}}After going into halftime down by eight , the Rockets\\ came out firing in the third quarter and out - scored\\ the Nuggets 59 - 42 to seal the victory on the road\end{tabular}} \\

\end{tabular}
\begin{tabular}{p{8.5cm}}
The \textbf{Houston Rockets} (\textbf{18}-\textbf{5}) defeated the \textbf{Denver Nuggets} (\textbf{10}-\textbf{13}) \textbf{108}-\textbf{96} on Saturday. \hlyellow{\textbf{Houston} has won}\red{\hlyellow{ \textbf{2 straight games} }}\hlyellow{and}\red{\hlyellow{ \textbf{6} }}\hlyellow{of their last}\red{\hlyellow{ \textbf{7}}}.  \hlyellow{\textbf{Dwight Howard} returned to action Saturday after missing the Rockets ' last}\red{\hlyellow{ \textbf{11 games} }}\hlyellow{with a knee injury}.  He was supposed to be limited to \red{\textbf{24 minutes}} in the game, but \blue{\textbf{Dwight Howard}} persevered to play \textbf{30 minutes} and put up a monstrous double-double of \blue{\textbf{26 points}} and \textbf{13 rebounds}.  Joining \textbf{Dwight Howard} in on the fun was \textbf{James Harden} with a triple-double of \textbf{24 points}, \textbf{10 rebounds} and \textbf{10 assists} in \textbf{38 minutes}. The \textbf{Rockets} ' formidable defense held the \textbf{Nuggets} to just \textbf{38 percent} shooting from the field. \hlyellow{\textbf{Houston} will face the \textbf{Nuggets} again in their next game, going on the road to \textbf{Denver} for their game on Wednesday}. \hlyellow{\textbf{Denver} has lost}\red{\hlyellow{ \textbf{4} }}\hlyellow{of their last}\red{\hlyellow{ \textbf{5 games} }}\hlyellow{as they struggle to find footing during a tough part of their schedule} ... \hlyellow{\textbf{Denver} will begin a}\red{\hlyellow{ \textbf{4 - game} }}\hlyellow{homestead hosting the San Antonio Spurs on Sunday}.
\end{tabular}
\caption{An example from the {\tt RotoWire} corpus. Partial box- and line-score tables are on the top left. Grounded entities and numerical facts are in \textbf{bold}. \hlyellow{Yellow} sentences contain \red{\textbf{red}} ungrounded numerical facts, and team game schedule related statements. A system-generated statement with multiple hallucinations on the bottom left.}
\label{tab:example}
\vspace{-4mm}
\end{table*}

Specifically, we observe that about 40\% of the game summary contents cannot be directly mapped to any input boxscore records, as exemplified by \autoref{tab:example}.
Written by professional sports journalists, these statements incorporate domain expertise and background knowledge consolidated from heterogeneous sources that are often hard to trace. The resulting information imbalance hinders a model to produce texts fully conditioned on the inputs and the uncontrolled randomness causes factual hallucinations, especially for the modern encoder-decoder framework~\citep{SutskeverVL14, ChoMBB14}. However, data fidelity is crucial for data-to-text generation besides fluency. In this real-world application, mistaken statements are detrimental to the document quality no matter how human-like they appear to be.

Apart from the popular BLEU~\citep{bleu} metric for text generation,~\citet{ws17} also formalized a set of post-hoc information extraction (IE) based evaluations to assess the data fidelity. Using the boxscore table schema, a sequence of (\textit{entity}, \textit{value}, \textit{type}) records mentioned in a system-generated summary are extracted as the content plan. They are then validated for accuracy against the boxscore table and similarity with the one extracted from the human-written summary. However, any hallucinated facts may unrealistically boost the BLEU score while not penalized by the data fidelity metrics since no records can be identified from the ungrounded contents. Thus the possibly misleading evaluation results inhibit systems to demonstrate excellence on this task.

These two aspects potentially undermine people's interests in this data fidelity oriented table-to-text generation task. 
Therefore, in this work, we revamp the task emphasizing this core aspect to further enable research in this direction.
First, we restore the information balance by trimming the summaries of ungrounded contents and replenish the boxscore table to compensate for missing inputs. 
This requires the non-trivial extraction of the latent gold standard content plans with high-quality.
Thus, we take the efforts to design sophisticated heuristics and achieved an estimated 98\% precision and 95\% recall of the true content plans, retaining 74\% of numerical words in the summaries. 
This yields better content plans as compared to the 94\% precision, 80\% recall by~\citet{ratishacl} and 60\% retainment by~\citet{ws17} respectively.
Guided by the high-quality content plans, only fact-grounded contents are identified and retained as shown in \autoref{tab:example}. 
Furthermore, by expending with 50\% more games between the years 2017-19, we obtain the more focused {\tt RotoWire-FG} ({\tt RW-FG}) dataset. 

This leads to more accurate evaluations and collectively paves the way for future works by providing a more user-friendly alternative.
With this refurbished setup, the existing models are then re-assessed on their abilities to ensure data fidelity. 
We discover that by only purifying the {\tt RW} dataset, the models can generate more precise facts without sacrificing fluency. 
Furthermore, we propose a new form of table reconstruction as an auxiliary task to improve fact grounding. By incorporating it into the state-of-the-art Neural Content Planning (NCP)~\citep{ratishaaai} model, we established a benchmark on the {\tt RW-FG} dataset with a 24.41 BLEU score and 95.7\% factual accuracy.

Finally, these insights lead us to summarize several fine-grained future challenges based on concrete examples, regarding factual accuracy and intra- and inter- sentence coherence.

Our contributions include:
\begin{enumerate}[topsep=3pt, itemsep=3pt, partopsep=3pt, parsep=3pt]
\item We introduce a purified, enlarged and enriched new dataset to support the more focused fact-grounded table-to-text generation task. We provide high-quality summary facts to table records mappings (content plan) and a more user-friendly experimental setup. All codes and data are freely available\footnote{\url{https://github.com/wanghm92/rw_fg}}.
\item We re-investigate existing methods with more insights, establish a new benchmark on this task, and uncover more fine-grained challenges to encourage future research.
\end{enumerate}

\section{Data-to-Text Dataset}
\label{sec:data}


This task requires models to take as inputs the NBA basketball game boxscore tables containing hundreds of records and generate the corresponding game summaries. 
A table can be view as a set of (\textit{entity}, \textit{value}, \textit{type}) records where \textit{entity} is the row name and \textit{type} is the column name in~\autoref{tab:example}.

\noindent\textbf{Formally}: Let $\mathbb{E} = \left\{e_{k}\right\}_{k=1}^{K}$ be the set of entities for a game. $\mathbb{S}=\left\{r_{j}\right\}_{j=1}^{S}$ be the set of records where each $r_j$ has a value $r_j^m$, an entity name $r_j^e$, a record type $r_j^t$ and $r_j^h$ indicating if the entity is the HOME or AWAY team. For example, a record has $r_j^t$ = POINTS, $r_j^e$ = Dwight Howard, $r_j^m$ = 26, and $r_j^h$ = HOME. The summary has $T$ words: $\hat{y}_{1 : T}=\hat{y}_{1}, \dots, \hat{y}_{T}$. A sample is a ($\mathbb{S}$, $\hat{y}_{1 : T}$) pair. 

\subsection{{\bf Looking into the }{\tt \textbf{RotoWire}} Corpus}
\label{sec:data:roto}
To better understand what kind of ungrounded contents are causing the interference, we manually examine a set of 30 randomly picked samples\footnote{For convenience, they are from the validation set and also used later for evaluation purposes.} and categorize the sentences into 5 types whose counts and percentages are tabulated in~\autoref{tab:rw}.

\begin{table}[]
\centering
\small
\begin{tabular}{l|c|c|c|c|c}
\hline
Type & His    & Sch    & Agg   & Game    & Inf    \\\hline\hline
Count   & 69   & 33   & 9   & 23   & 23   \\\hline
Percent   & 43.9 & 21.0 & 5.7 & 14.7 & 14.7 \\\hline
\end{tabular}
\caption{Types of ungrounded contents about statistics related to \underline{\textit{His}}: history (e.g. recent-game/career high/average) \underline{\textit{Sch}}: team schedule (e.g. what is next game); \underline{\textit{Agg}}: aggregation of statistics from multiple players (e.g. the duo of two stars combined scoring ...) ; \underline{\textit{Game}}: during the game (e.g. a game winning shot with 1 second left); \underline{\textit{Inf}}: inferred from aggregations (e.g. a player carried the team for winning)}
\label{tab:rw}
\end{table}

The \underline{\textit{His}} type occupies the majority portion, followed by the game-specific \underline{\textit{Game}}, \underline{\textit{Inf}}, and \underline{\textit{Agg}} types, and the remaining goes to \underline{\textit{Sch}}.
Specifically, the \underline{\textit{His}} and \underline{\textit{Agg}} types come from exponentially large number of possible combinations of game statistics, and the \underline{\textit{Inf}} type is based on subjective judgments. Thus, it is difficult to trace and aggregate the heterogeneous sources of origin for such statements to fully balance the input and output. The \underline{\textit{Sch}} and \underline{\textit{Game}} types require a sample from a large pool of non-numerical and time-related information, whose exclusion would not affect the nature of the fact-grounding generation task. 
On the other hand, these ungrounded contents misguide a system to generate hallucinated facts and thus defeat the purpose of developing and evaluating models for fact-grounded table-to-text generation. 
Thus, we emphasize on this core aspect of the task by trimming contents not licensed by the boxscore table, which we show later still encompasses many fine-grained challenges awaiting to be resolved.
While fully restoring all desired inputs is also an interesting research challenge, it is orthogonal to our focus and thus left for future explorations.

\subsection{{\tt \textbf{RotoWire-FG}}}
\label{sec:data:fg}
Motivated by these observations, we perform purification and augmentation on the original dataset to obtain the new {\tt RW-FG} dataset. 

\subsubsection{Dataset Purification}
\label{sec:data:fg:pur}

\noindent\textbf{Purifying Contents}: 
We aim to retain game summary contents with facts licensed by the boxscore records. 
The sports game summary genre is more descriptive than analytical and aims to concisely cover salient player or team statistics. 
Correspondingly, a summary often finishes describing one entity before shifting to the next. 
This fashion of topic shift allows us to identify the topic boundaries using sentences as units, and thus greatly narrows down the candidate boxscore records to be aligned with a fact. 
The mappings can then be identified using simple pattern-based matching, as also explored by~\citet{ws17}. 
It also enables resolving co-reference by mapping the singular and plural pronouns to the most recently mentioned players and teams respectively.
A numerical value associated with an entity is licensed by the boxscore table if it equals to the record value of the desired type.
Thus we design a set of heuristics to determine the types, such as mapping ``Channing Frye furnished 12 points'' to the (\textit{Channing Frye}, \textit{12}, \textit{POINTS}) record in the table.
Finally, consecutive sentences describing the same entity is retained if any numerical value is licensed by the boxscore table. 

This trimming process introduces negligible influences on the inter-sentence coherence for the summaries. 
We achieve a 98\% precision and a 95\% recall of the true content plans and align 74\% of all numerical words in the summaries to records in the boxscore tables.
The sequence of mapped records is extracted as the content plans and samples describing fewer than 5 records are discarded. 

In between the labor-intensive yet imperfect manual annotation and the cheap but inaccurate lexical matching, we achieved better quality through designing the heuristics using similar efforts as training and assembling the IE models by~\citet{ws17}. Meanwhile, more accurate content plans provide better reliability during evaluation.

\noindent\textbf{Normalization}: To enhance accuracy, we convert all English number words into numerical values. As some percentages are rounded differently between the summaries and the boxscore tables, such discrepancies are rectified. We also perform entity normalization for players and teams, resolving mentions of the same entity to one lexical form. This makes evaluations more user-friendly and less prone to errors.

\subsubsection{Dataset Augmentation}
\label{sec:data:fg:aug}

\begin{table}[t]
\small
\centering
\setlength{\tabcolsep}{2pt}
\begin{tabular}{lccccc}
\hline
Versions & Examples & Tokens & Vocab & Types & Avg Len \\ \hline\hline
{\tt RW} & 4.9K & 1.6M & 11.3K & 39 & 337.1 \\ 
{\tt RW-EX} & 7.5K & 2.5M & 12.7K & 39 & 334.3 \\\hline
{\tt RW-FG} & 7.5K & 1.5M & 8.8K & 61 & 205.9 \\\hline
\end{tabular}
\caption{Comparison between datasets. ({\tt RW-EX} is the enlarged {\tt RW} with 50\% more games)}
\hspace{10em}
\begin{tabular}{c|c|c|c|c|}
\cline{2-5}
&Sents & Content Plans & Records & Num-only Records \\ \hline
\multicolumn{1}{|l|}{\tt RW-EX} & 14.0 & 27.2 & 494.2 & 429.3 \\ \hline
\multicolumn{1}{|l|}{\tt RW-FG} & 8.6 & 28.5 & 519.9 & 478.3 \\ \hline
\end{tabular}
\caption{Dataset statistics by the average number of each item per sample.}
\label{tab:dataset}
\end{table}

\noindent\textbf{Enlargement}: 
Similar to~\citet{ws17}, we crawl the game summaries from the \textit{RotoWire Game Recaps}\footnote{\url{https://www.RotoWire.com/basketball/game-recaps.php}} between years 2017-19 and align the summaries with the official \textit{NBA}\footnote{\url{https://stats.nba.com/}} boxscore tables. This brings 2.6K more games with 56\% more tokens, as tabulated in \autoref{tab:dataset}.

\noindent\textbf{Line-score replenishment}:
Many team statistics in the summaries are missing in the line-score tables. We recover them by aggregating other boxscore statistics. 
For example, the number of shots attempted and made by the team for field goals, 3-pointers, and free-throws are calculated by summing their player statistics. Besides, we supplement a set of team point breakdowns as shown in \autoref{tab:linescore}. The replenishment boosts the recall on numerical values from 72\% to 74\% and augments the content plans by 1.3 records per sample.

\setlength{\aboverulesep}{0pt}
\setlength{\belowrulesep}{0pt}

\begin{table}[]
\centering
\small
\setlength{\tabcolsep}{4pt}
\begin{tabular}{c|c|c|c|c|c|c|}
\cmidrule{2-7}
 & \multicolumn{4}{c|}{\cellcolor[HTML]{FFFC9E}Quarters} & \multicolumn{2}{c|}{\cellcolor[HTML]{FFFC9E}Players} \\ \hline
\rowcolor[HTML]{CBCEFB} 
\multicolumn{1}{|c|}{\cellcolor[HTML]{9AFF99}Sums} & 1 to 2 & 1 to 3 & 2 to 3 & 2 to 4 & bench & starters \\ \hline
 & \multicolumn{2}{c|}{\cellcolor[HTML]{FFFC9E}Halves} & \multicolumn{4}{c|}{\cellcolor[HTML]{FFFC9E}Quarters} \\ \hline
\rowcolor[HTML]{CBCEFB} 
\multicolumn{1}{|c|}{\cellcolor[HTML]{9AFF99}Diffs} & 1st & 2nd & 1 & 2 & 3 & 4 \\ \hline
\end{tabular}
\caption{Replenished line-score statistics. Each purple cell corresponds to a new record type, defined as applying the the operation in the row names (green) to the source of statistics in the column names (yellow). ``Sums'' operates on individual teams and ``Diffs'' is between the two teams. For example, the ``1 to 2'' cell in the second row means the summation of points scored by a team in the 1st and 2nd ``Quarters'', the ``1st'' cell in the fourth row means the difference between the two teams' 1st half points. }
\label{tab:linescore}
\end{table}

\noindent\textbf{Finalize}: We conduct the same purification procedures described in \autoref{sec:data:fg:pur} after the augmentations. More data collection details are included in~\autoref{sec:appendix}.

\section{Re-assessing Models on Purified {\tt \textbf{RW}}}
\label{sec:rw}

\subsection{Models}
\label{sec:rw:model}
We re-assess three neural network based models on this task\footnote{\citet{iso-etal-2019-learning} was released after this work was submitted. It also altered the {\tt RW-FG} dataset for experiments, so the results would not be directly comparable. The method is worth investigation for future works.}. To feed the tables to the models, each record $r_j$ has attribute embeddings for $r_j^m$, $r_j^e$, $r_j^t$, $r_j^h$ and their concatenation is the input.
\begin{itemize}
    \item ED-CC~\citep{ws17}: This is an Encoder-Decoder (ED)~\cite{SutskeverVL14, ChoMBB14} model with an 1-layer MLP encoder~\citep{yang17}, and an LSTM~\citep{lstm} decoder with the Conditional Copy (CC) mechanism~\citep{GulcehreANZB16}.
    
    \item NCP~\citep{ratishaaai}: The Neural Content Planning (NCP) model employs a pointer network~\citep{VinyalsFJ15} to select a subset of records from the boxscore table and sequentially roll them out as the content plan. Then the summary is then generated only from the content plan using the ED-CC model with a Bi-LSTM encoder. 
    
    \item ENT~\citep{ratishacl}: The ENTity memory network (ENT) model extends the ED-CC model with a dynamically updated entity-specific memory module to capture topic shifts in outputs and incorporate it into each decoder step with a hierarchical attention mechanism.
\end{itemize}

\subsection{Evaluation}
\label{sec:rw:eval}
In addition to using BLEU~\citep{bleu} as a reasonable proxy for evaluating the fluency of the generated summary,~\citet{ws17} designed three types of metrics to assess if a summary accurately conveys the desired information.

\noindent\textbf{Extractive Metrics}: First, an ordered sequence of (\textit{entity}, \textit{value}, \textit{type}) triples are extracted from the system output summary as the content plan using the same heuristics in \autoref{sec:data:fg:pur}. It is then checked against the table for its accuracy (RG) and the gold content plan to measure how well they match (CS \& CO). 
Specifically, let $cp=\{r_i\}$ and $cp'=\{{r'}_i\}$ be the gold and system content plan respectively, and $\left | . \right |$ denote set cardinality.
We calculate the following measures: 

\begin{itemize}

\item \textbf{Content Selection (CS)}: 

\begin{itemize}
    \item Precision (\textbf{CSP}) = $\left | cp \cap cp' \right |/\left | cp' \right|$
    \item Recall (\textbf{CSR}) = $\left | cp \cap cp' \right |/\left | cp \right|$
    \item F1 (\textbf{CSF}) = $2PR/(P+R)$
\end{itemize}

\item \textbf{Relation Generation (RG)}: 

\begin{itemize}
    \item Count(\textbf{\#}) =  $\left | cp' \right |$
    \item Precision (\textbf{RGP}) = $\left | cp' \cap \mathbb{S} \right |/\left | cp' \right|$
\end{itemize}

\item \textbf{Content Ordering (CO)}:

\begin{itemize}
    \item \textbf{DLD}: normalized Damerau Levenshtein Distance~\citep{dld} between $cp$ and $cp'$
\end{itemize}

\end{itemize}

\textbf{CS} and \textbf{RG} measures the ``what to say'' and \textbf{CO} measures the ``how to say'' aspects.

\subsection{Experiments}
\label{sec:rw:exp}

\noindent\textbf{Setup}: 
To re-investigate the existing three methods on the ability to convey accurate information conditioned on the input, we assess them by training on the purified {\tt RW} corpus.
To demonstrate the differences brought by the purification process, we keep all other settings unchanged and report results on the original validation and test sets after performing early stopping~\citep{yao2007early} based on the BLEU score.

\begin{savenotes}
\begin{table*}[ht]
\centering
\small
\setlength{\tabcolsep}{4pt}
\begin{tabular}{l|cc|ccc|c|cc|ccc|c}
\hline
\multirow{3}{*}{Model} & \multicolumn{6}{c|}{Dev} & \multicolumn{6}{c}{Test} \\
\cline{2-13}
 & \multicolumn{2}{c|}{RG} & \multicolumn{3}{c|}{CS} & CO & \multicolumn{2}{c|}{RG} & \multicolumn{3}{c|}{CS} & CO \\
 & \# & P\% & P\% & R\% & F1\% & DLD\% & \# & P\% & P\% & R\% & F1\% & DLD\% \\
 \hline
ED-CC & 23.95 & 75.10 & 28.11 & 35.86 & 31.52 & 15.33 & 23.72 & 74.80 & 29.49 & 36.18 & 32.49 & 15.42 \\
ED-CC({\tt FG}) & 22.65 & 78.63 & 29.48 & 34.08 & 31.61 & 14.58 & 23.36 & 79.88 & 29.36 & 33.36 & 31.23 & 13.87 \\\hline
NCP & 33.88 & 87.51 & 33.52 & 51.21 & 40.52 & 18.57 & 34.28 & 87.47 & 34.18 & 51.22 & 41.00 & 18.58 \\
NCP({\tt FG}) & 31.90 & 90.20 & 34.53 & 49.74 & 40.76 & 18.29 & 33.51 & 91.46 & 33.96 & 49.14 & 40.16 & 18.16 \\\hline
ENT\footnote{For fair comparison, we report results of ENT model after fixing a bug in the evaluation script as endorsed by the author of~\citet{ws17} at \url{https://github.com/harvardnlp/data2text/issues/6}} & 21.49 & 91.17 & 40.50 & 37.78 & 39.09 & 19.10 & 21.53 & 91.87 & 42.61 & 38.31 & 40.34 & 19.50 \\
ENT({\tt FG}) & 30.08 & 93.74 & 30.43 & 48.64 & 37.44 & 16.53 & 30.66 & 93.09 & 32.40 & 41.69 & 36.46 & 16.44 \\ \hline
\end{tabular}
\caption{Comparison between models trained on {\tt RW} and {\tt RW-FG}}
\label{tab:rw:non-bleu}
\end{table*}
\end{savenotes}

\begin{table*}[thb]
\centering
\small
\setlength{\tabcolsep}{4pt}
\begin{tabular}{l|cccccc|cccccc}
\hline
\multirow{2}{*}{Model} & \multicolumn{6}{c|}{Dev} & \multicolumn{6}{c}{Test} \\
\cline{2-13}
              & B1 & B2 & B3 & B4 & BP & BLEU & B1 & B2 & B3 & B4 & BP & BLEU \\
\hline
ED-CC & 44.42 & \textbf{18.16} & \textbf{9.40} & 5.95 & \textbf{1.00} & \textbf{14.57} & 43.22 & \textbf{17.64} & \textbf{9.16} & 5.81 & \textbf{1.00} & \textbf{14.19} \\
ED-CC({\tt FG}) & \textbf{46.61} & 17.70 & 9.33 & \textbf{6.21} & 0.59 & 8.74 & \textbf{45.75} & 17.14 & 9.05 & \textbf{5.98} & 0.61 & 8.68 \\ \hline
NCP & 48.95 & 20.58 & 10.70 & 6.96 & \textbf{1.00} & \textbf{16.19} & 49.77 & 21.19 & 11.31 & 7.46 & \textbf{0.96} & \textbf{16.50} \\
NCP({\tt FG}) & \textbf{56.63} & \textbf{24.15} & \textbf{12.45} & \textbf{8.13} & 0.54 & 10.45 & \textbf{56.33} & \textbf{23.92} & \textbf{12.42} & \textbf{8.11} & 0.53 & 10.25 \\ \hline
ENT & 51.57 & 21.92 & 11.87 & 8.08 & \textbf{0.88} & \textbf{15.97} & 53.23 & \textbf{23.07} & \textbf{12.78} & \textbf{8.78} & \textbf{0.84} & \textbf{16.12} \\
ENT({\tt FG}) & \textbf{56.08} & \textbf{23.29} & \textbf{12.29} & \textbf{8.16} & 0.44 & 8.92 & \textbf{55.03} & 21.86 & 11.38 & 7.38 & 0.57 & 10.17 \\ \hline
\end{tabular}
\caption{Breakdown of BLEU scores for models trained on {\tt RW} and {\tt RW-FG}}
\label{tab:rw:bleu}
\end{table*}

\noindent\textbf{Results}: 
As shown in \autoref{tab:rw:non-bleu}, we observe increase in Relation Generation Precision (RGP) and on-par performance for Content Selection (CS) and Content Ordering (CO). In particular, Relation Generation Precision (RGP) is substantially increased by an average 2.7\% for all models. The Content Selection (CS) and Content Ordering (CO) measures fluctuate above and below the references, with the biggest disparity on Content Selection Precision (CSP), Content Selection Recall (CSR) and Content Ordering (CO) for the ENT model. Since output length is a main independent variable for this set of experiments and a crucial factor in BLEU score as well, we report the breakdowns in~\autoref{tab:rw:bleu}. Specifically, the NCP model shows consistent improvements on all BLEU 1-4 scores, similarly for ENT on the validation set. Among all fluctuation around the references, nearly all models demonstrate an increase in BLEU-1 and BLEU-4 precision. Reflected on the BP coefficients, models trained on the purified summaries produces shorter outputs, which is the major reason for lower BLEU scores when using the un-purified summaries as the references.


\subsection{How Purification Affects Performance}
\label{sec:rw:dis}

First, simply replacing with the purified training set leads to considerable improvements in the Relation Generation Precision (RGP). This is because removing the ungrounded facts (e.g. \underline{\textit{His}}, \underline{\textit{Agg}}, and \underline{\textit{Game}} types) alleviates their interference with the model while learning when and where to copy over a correct numerical value from the table. 
Besides, since the ungrounded facts do not contribute to the gold or system output content plan during the information extraction process, the other extractive metrics Content Selection (CS) and Content Ordering (CO) measures stay on-par.

One abnormality is the big difference in the Content Selection (CS) and Content Ordering (CO) measures from the ENT model. This is not that surprising after examining the outputs, which appear to collapse into template-like summaries. For example, 97.8\% sentences start with the game points followed by a pattern ``\underline{XX} were the superior shooters'' where \underline{XX} represents a team. Tracing back to the model design, it is explicitly trained to model topic shifts on the token level during generation, which instead happens more often on the sentence level. As a result, it degenerates to remembering a frequent discourse-level pattern from the training data. We observe a similar pattern on the outputs from original outputs by~\citet{ratishacl}, which is aggravated when trained on the purified dataset. On the other hand, the NCP model decouples the content selection and planning on the discourse level from the surface realization on the token level, and thus generalizes better.

\section{A New Benchmark on {\tt \textbf{RW-FG}}}
\label{sec:new}
With more insights about the existing methods, we take a step further to achieve better data fidelity.
~\citet{ws17} achieved improvements on the ED with Joint Copy (JC)~\cite{gu16} model by introducing an reconstruction loss~\citep{TuLSLL17} during training. 
Specifically, the decoder states at each time step are used to predict record values in the table to enable broader input information coverage.

However, we take a different point of view: one key mechanism to avoid reference errors is to ensure that the set of numerical values mentioned in a sentence belongs to the correct entity with the correct record field type. 
While the ED-CC model is trained to achieve such alignments, it should also be able to accurately fill the numbers back to the correct cells in an empty table. 
This should be done by only accessing the column and row information of the cells without explicitly knowing the original cell values. 
Further leveraging on the planner output of the NCP model, the candidate cells to be filled can be reduced to the content plan cells selected by the planner. 
With this intuition, we devise a new form of table reconstruction (TR) task incorporated into the NCP model. 

Specifically, each content plan record has attribute embeddings for $r_j^e$, $r_j^t$, and $r_j^h$, excluding its value, and we encode them using a 1-layer MLP~\citep{yang17}. We then employ the~\citet{LuongPM15} attention mechanism at each $\hat{y}_{t}$ if it is a numerical value with the encoded content plan as the memory bank. The attention weights are then viewed as probabilities of selecting each cell to fill the number $\hat{y}_{t}$. The model is additionally trained to minimize the negative log-likelihood of the correct cell.

\subsection{Experiments}
\label{sec:new:exp}
\noindent\textbf{Setup}: 
We assess models on the {\tt RW-FG} corpus to establish a new benchmark. Following~\citet{ws17}, we split all samples into train (70\%), validation (15\%), and test (15\%) sets, and perform early stopping~\citep{yao2007early} using BLEU~\citep{bleu}. We adapt the template-based generator by~\citet{ws17} and remove the ungrounded end sentence since they are eliminated in {\tt RW-FG}.

\begin{table*}[thb!]
\centering
\small
\setlength{\tabcolsep}{3.65pt}
\begin{tabular}{l|cc|ccc|c|c|cc|ccc|c|c}
\hline
\multirow{3}{*}{Model} & \multicolumn{7}{c|}{Dev} & \multicolumn{7}{c}{Test} \\ \cline{2-15} 
 & \multicolumn{2}{c|}{RG} & \multicolumn{3}{c|}{CS} & CO & \multirow{2}{*}{BLEU} & \multicolumn{2}{c|}{RG} & \multicolumn{3}{c|}{CS} & CO & \multirow{2}{*}{BLEU} \\

 & \# & P\% & P\% & R\% & F1\% & DLD\% &  & \# & P\% & P\% & R\% & F1\% & DLD\% &  \\ \hline
TMPL & 51.81 & 99.09 & 23.78 & 43.75 & 30.81 & 10.06 & 11.91 & 51.80 & 98.89 & 23.98 & 43.96 & 31.03 & 10.25 & 12.09 \\
WS17 & 30.47 & 81.51 & 36.15 & 39.12 & 37.57 & 18.56 & 21.31 & 30.28 & 82.16 & 35.84 & 38.40 & 37.08 & 18.45 & 20.80 \\
ENT & 35.56 & 93.30 & 40.19 & 50.71 & 44.84 & 17.81 & 21.67 & 35.69 & 93.72 & 39.04 & 49.29 & 43.57 & 17.50 & 21.23 \\
NCP & 36.28 & 94.27 & \textbf{43.31} & 55.96 & 48.91 & 24.08 & 24.49 & 35.99 & 94.21 & \textbf{43.31} & 55.15 & 48.52 & 23.46 & 23.86 \\ \hline
NCP+TR & \textbf{37.04} & \textbf{95.65} & 43.09 & \textbf{57.24} & \textbf{49.17} & \textbf{24.75} & \textbf{24.80} & \textbf{37.49} & \textbf{95.70} & 42.90 & \textbf{56.91} & \textbf{48.92} & \textbf{24.47} & \textbf{24.41} \\ \hline
\end{tabular}
\caption{Performances of models on {\tt RW-FG}}
\label{tab:new:exp}
\end{table*}

\noindent\textbf{Results}: 
As shown in \autoref{tab:new:exp}, the template model can ensure high Relation Generation Precision (RGP) but is inflexible as shown by other measures. Different from~\citet{ratishacl}, the NCP model is superior on all measures among the baseline neural models. The ENT model only outperforms the basic ED-CC model but surprisingly yields lower Content Selection (CS) measures. Our NCP+TR model outperforms all baselines except for slightly lower Content Selection Precision (CSP) compared to the NCP model.

\subsection{Discussion}
\label{sec:new:dis}

We observe that the ED-CC model produces the least number of candidate records, and correspondingly achieves the lowest Content Selection Recall (CSR) compared to the gold standard content plans.
As discussed in \autoref{sec:rw:dis}, the template-like discourse pattern produced by the ENT model noticeably deteriorates its performance. It is completely outperformed by the NCP model and even achieves lower CO-DLD than the ED-CC model.
Finally, as supported by the extractive evaluation metrics, employing table reconstruction as an auxiliary task indeed boosts the decoder to produce more accurate factual statements.
We discuss in more detail as follows.

\subsubsection{Manual Evaluation}
\label{sec:new:dis:manual}

\begin{table}[]
\centering
\small
\setlength{\tabcolsep}{4pt}
\begin{tabular}{l|c|c|c|c|c}
\hline
Model & Total(\#) & RP(\%) & WC(\%) & UG(\%)   & IC(\%) \\ \hline\hline
NCP   & 246   & 9.21   & 11.84 & 3.07 & 5.26       \\ \hline
NCP+TR & 228   & 3.66   & 8.94  & 3.25 & 2.03       \\ \hline
\end{tabular}
\caption{Error types of manual evaluation. \underline{\textit{Total}}: number of sentences; \underline{\textit{RP}}: Repetition; \underline{\textit{WC}}: Wrong Claim; \underline{\textit{UG}}: Ungrounded sentence; \underline{\textit{IC}}: Incoherent sentence}.
\label{tab:new:man}
\end{table}

To gain more insights into how exactly NCP+TR improves from NCP in terms of factual accuracy, we manually examined the outputs on the 30 samples. We compare the two systems after categorizing the errors into 4 types. As shown in \autoref{tab:new:man}, the largest improvement comes from reducing repeated statements and wrong fact claims, where the latter involves referring to the wrong entity or making the wrong judgment of the numerical value. The NCP+TR generally produces more concise outputs with a reduction in repetitions, consistent with the objective for table reconstruction.

\begin{table}[ht]
\centering
\small
\begin{tabular}{|p{7.3cm}|}
\hline
The \textbf{Cleveland Cavaliers} defeated the \textbf{Philadelphia 76ers} , \textbf{102} - \textbf{101} , at \textbf{Wells Fargo Center} on Monday evening . 
\textbf{LeBron James} led the way with a \textbf{25 - point} , \textbf{14 - assist} double double that also included \textbf{8 rebounds} , \textbf{2 steals} and \textbf{1 block} .
\textbf{Kevin Love} followed with a \textbf{20 - point} , \textbf{11 - rebound} double double that also included \textbf{1 assist} and \textbf{1 block} . 
\textbf{Channing Frye} led the bench with \textbf{12 points} , \textbf{2 rebounds} , \textbf{2 assists} and \textbf{2 steals}
\textbf{Kyrie Irving} managed \textbf{8 points} , \textbf{7 rebounds} , \textbf{2 assists} and \textbf{2 steals} . 
...
\hlmelon{\textbf{Joel Embiid} 's \textbf{22 points} led the \textbf{Sixers} , a total he supplemented with \textbf{6 rebounds} , \textbf{2 assists} , \textbf{4 blocks} and \textbf{1 steal}}
...
\\\hline
The \textbf{Cleveland Cavaliers} defeated the \textbf{Philadelphia 76ers} , \textbf{102} - \textbf{101} , at \textbf{Wells Fargo Center} on Friday evening . 
\hlgreen{The \textbf{Cavaliers} came out of the gates hot , jumping out to a \textbf{34 - 15} lead after 1 quarter . However , the \textbf{Sixers} ( \textbf{0} - \textbf{5} ) stormed back in the second to cut the deficit to just \textbf{2} points by halftime . However , the light went on for Cleveland at intermission , as they built a \textbf{9} - point lead by halftime} .
\textbf{LeBron James} led the way for the Cavaliers with a \textbf{25 - point} , \textbf{14 - assist} double double that also included \textbf{8 rebounds} , \textbf{2 steals} and \textbf{1 block} .
\hlcyan{Kyrie Irving followed \textbf{Kevin Love} with a \textbf{20 - point} , \textbf{11 - rebound} double double that also included \textbf{1 assist} and \textbf{1 block}} . 
\hlyellow{\textbf{Channing Frye} furnished \textbf{12 points} , \textbf{2 rebounds} , \textbf{2 assists} and \textbf{2 steals}}
...
\hlyellow{\textbf{Channing Frye} led the bench with \textbf{12 points} , \textbf{2 rebounds} , \textbf{2 assists} and \textbf{2 steals}} . 
\hlmelon{\textbf{Jahlil Okafor} led the \textbf{Sixers} with \textbf{22 points}, \textbf{6 rebounds} , \textbf{2 assists}, \textbf{4 blocks} and \textbf{1 steal}}
...
\textbf{Jahlil Okafor} managed \textbf{14 points} , \textbf{5 rebounds} , \textbf{3 blocks} and \textbf{1 steal} .\\\hline
\end{tabular}
\caption{Case study comparing NCP+TR (above) and NCP (below). The records identified are in \textbf{bold}. The pair of sentences in \hlmelon{orange} shows an referring error to \textbf{Jahlil Okafor} is corrected above to \textbf{Joel Embiid}, where all the trailing statistics actually belong to \textbf{Joel Embiid}, and \textbf{Jahlil Okafor}'s actual statistics are described at the end. The \hlyellow{yellow} sentences repeats on the same player. The \hlgreen{green} sentences actually shows some more contents selected by the NCP model. The \hlcyan{blue} sentence is a tricky one, where it should describe \textbf{Kyrie Irving}'s statistics but actually describing \textbf{Kevin Love}'s but the summary above does not have this issue.}
\label{tab:new:case}
\end{table}

\subsubsection{Case study}
\label{sec:new:dis:case}
\autoref{tab:new:case} shows a pair of outputs by the two systems. In this example, the NCP+TR model can correct wrong the player name ``\textit{Jahlil Okafor}'' by ``\textit{Joel Embiid}'', while keeping the statistics intact. It also avoids repeating on ``\textit{Channing Frye}'' and the semantically incoherent expression about ``\textit{Kevin Love}'' and ``\textit{Kyrie Irving}''. Nonetheless, this NCP output selects more records to describe the progress of the game. This shows how the NCP+TR trained with more constraints behaves more accurately but conservatively.

\section{Errors and Challenges}
\label{sec:challenage}

\begin{table}[ht]
\centering
\small
\begin{tabular}{p{7.3cm}}
\noindent\blue{\textbf{(1) Intra-sentence coherence}}: 
\begin{itemize}[leftmargin=*,topsep=3pt,itemsep=0pt,partopsep=0pt, parsep=3pt,label=\raisebox{0.25ex}{\tiny$\bullet$}]
    \itemsep0em
    \item The \textbf{Lakers} were the \red{\textit{superior}} \red{\textit{shooters}} in this game , going \textbf{48 percent} from the field and \red{\textbf{24 percent}} from the three point line , while the \textbf{Jazz} went \textbf{47 percent} from the floor and just \red{\textbf{30 percent}} from beyond the arc.
    \item The \textbf{Rockets} got off to a quick start in this game, \red{\textit{out scoring}} the \textbf{Nuggets} \red{\textbf{21}}-\red{\textbf{31}} right away in the 1st quarter.
\end{itemize}
\\
\blue{\textbf{(2) Inter-sentence coherence}}: 

\begin{itemize}[leftmargin=*,topsep=3pt,itemsep=0pt,partopsep=0pt, parsep=3pt,label=\raisebox{0.25ex}{\tiny$\bullet$}]
    \itemsep0em
    \item \textbf{LeBron James} was the lone bright spot for the \textbf{Cavaliers} , as he led the team with \textbf{20 points} . \textbf{Kevin Love} was \red{\textit{the only Cleveland starter in double figures}} , as he tallied \textbf{17 points} , \textbf{11 rebounds} and \textbf{3 assists} in the loss. 

    \item  \red{\textit{\textbf{Dirk Nowitzki} led the \textbf{Mavericks} in scoring}} , finishing with \textbf{22 points} ( \textbf{7 - 13 FG} , \textbf{3 - 5 3PT} , \textbf{5 - 5 FT} ) , \textbf{5 rebounds} and \textbf{3 assists} in \textbf{37 minutes}\textbf{}. \red{\textit{He}} 's had a very strong stretch of games , scoring \textbf{17 points} on \textbf{6} - for - \textbf{13} shooting from the field and \textbf{5} - for - \textbf{10} from the three point line. \red{\textit{\textbf{JJ Barea} finished with \textbf{32 points}}} ( \textbf{13 - 21 FG} , \textbf{5 - 8 3PT} ) and \textbf{11 assists} ...
\end{itemize}

\\
\blue{\textbf{(3) Incorrect claim}}: 

\begin{itemize}[leftmargin=*,topsep=3pt,itemsep=0pt,partopsep=0pt, parsep=3pt, label=\raisebox{0.25ex}{\tiny$\bullet$}]
    \itemsep0em
    \item The \textbf{Heat} were able to force \red{\textbf{20}} turnovers from the \textbf{Sixers}, which may have been the difference in this game.
\end{itemize}
\\
\end{tabular}
\caption{Cases for three major types of system errors}
\label{tab:error}
\end{table}

Having revamped the task with better focus, re-assessed existing and improved models, we discuss 3 future directions in this task with concrete examples in \autoref{tab:error}: 
 
\noindent\textbf{Content Selection}: Since writers are subjective in choosing \textit{what to say} given the boxscore, it is unrealistic to force a model to mimic all kinds of styles. However, a model still needs to learn from training to select both the salient (e.g. surprisingly high/low statistics for a team/player) and the popular (e.g. the big stars) statistics.
One potential direction is to involve multiple human references to help reveal such saliency and make Content Ordering (CO) and Content Selection (CS) measures more interpretive. 
This is particularly applicable for the sports domain since a game can be uniquely identified by the teams and date but mapped to articles from different sources. Besides, multi-reference has been explored for evaluating data-to-text generation systems~\citep{NovikovaDR17} and for content selection and planning~\citep{GehrmannDER18}. It has also been studied in machine translation for evaluation~\citep{dreyer-marcu-2012-hyter} and training~\citep{ZhengM018}.

\noindent\textbf{Content Planning}: Content plans have been extracted by linearly rolling out the records and topic shifts are modeled as sequential changes between adjacent entities. However, this fashion does not reflect the hierarchical discourse structures of a document and thus ensures neither intra- nor inter-sentence coherence. As shown by the errors in (1) in \autoref{tab:error}, the links between entities and their numerical statistics are not strictly monotonic and switching the order results in errors.

On the other hand, autoregressive training for creating such content plans limits the model to capture frequent sequence patterns rather than allowing diverse arrangements.~\citet{moryossef19} demonstrates isolating the content planning from the joint end-to-end training and employing multiple valid content plans during testing. 
Although the content plan extraction heuristics are dataset-dependent, it is worth exploring for data in a closed domain like {\tt RW}.

\noindent\textbf{Surface Realization}: Although the NCP+TR model has achieved nearly 96\% Relation Generation Precision (RGP), it is still paramount to keep on improving data accuracy since one single mistake is destructive to the whole document. The challenge is more with the evaluation metrics. Specifically, all extractive metrics only validate if an extracted record maps to the true entity and type but disregards the semantics of its contexts. For example (2) in \autoref{tab:error}, even assuming the linear ordering of records, their context still causes inter-sentence incoherence. In particular, both LeBron and Kevin scored double digits and JJ Barea leads the scores rather than Dirk. For another example (3), the 20 turnovers records are selected to be Heat's but expressed falsely as Sixers'. As pointed out by~\citet{ws17}, this may require the integration of semantic or reference-based constraints during generation. The number magnitudes should be incorporated. For example,~\citet{nie-etal-2018-operation} has devised an interesting idea to implicitly improve coherence by supplementing the input with pre-computed results from algebraic operations on the table. Moreover,~\citet{qin-etal-2018-learning} proposed to automatically align the game summary with the record types in the input table on the phrase level. It can potentially be combined with the operation results to correct incoherence errors and improve the generations.


\section{Related Works}
\label{sec:rel}
Various forms of structured data has been used as input for data-to-text generation tasks, such as tree~\citep{Belz2011TheFS, Belz2018TheFM},  graph~\citep{Konstas2017NeuralAS}, dialog moves~\citep{NovikovaDR17},  knowledge base~\citep{Gardent2017TheWC, ChisholmRH17}, database~\citep{Konstas2017NeuralAS, gardent-etal-2017-webnlg, wang18}, and table~\citep{ws17,LebretGA16}. The {\tt RW} corpus we studied is from the sports domain which has attracted great interests~\citep{ChenM08, mei16, ratishacl}. However, unlike generating the one-entity descriptions~\citep{LebretGA16, wang18} or having the output strictly bounded by the inputs~\citep{NovikovaDR17}, this corpus poses additional challenges since the targets contain ungrounded contents. To facilitate better usage and evaluation of this task, we hope to provide a refined alternative, similar to the purpose by~\citet{enriching18}.


\section{Conclusion}
\label{sec:con}
In this work, we study the core fact-grounding aspect of the data-to-text generation task and contribute a purified, enlarged, and enriched {\tt RotoWire-FG} corpus with a more fair and reliable evaluation setup. We re-assess existing models and found that the more focused setting helps the models to express more accurate statements and alleviate fact hallucinations. Improving the state-of-the-art model and setting a benchmark on the new task, we reveal fine-grained unsolved challenges hoping to inspire more research in this direction.

\section*{Acknowledgments}
Thanks for the generous and valuable feedback from the reviewers.
Special thanks to Dr. Jing Huang and Dr. Yun Tang for their unselfish guidance and support.

\bibliography{acl2019}
\bibliographystyle{acl_natbib}

\appendix

\section{Appendices}
\label{sec:appendix}

\subsection{Data Collection Details}
\label{sec:appendix:data}
\begin{itemize}
    \item We use the text2num\footnote{\url{https://github.com/ghewgill/text2num/blob/master/text2num.py}} package to convert all English number words into numerical values
    \item We first get the summary title, date, and the contents from \textit{RotoWire Game Recaps}. The title contains the home and visiting team. Together with the date, this game is uniquely identified with a \textit{GAME\_ID}. Then we use the \textit{nba\_api}\footnote{\url{https://github.com/swar/nba_api}} package to query the \textit{stats.nba.com} by \textit{NBA.com}\footnote{\url{www.nba.com} ; \url{https://stats.nba.com/}} to obtain the game boxscore and line scores.~\citet{ws17} used the \textit{nba\_py}\footnote{\url{https://github.com/seemethere/nba_py}} package , which unfortunately has become obsolete due to lack of maintenance. To obtain the line scores with the same set of column types as the original RotoWire dataset, we collectively used two APIs, \textit{BoxScoreTraditionalV2} and \textit{BoxScoreSummaryV2}.
\end{itemize}

\end{document}